\documentclass[10pt,twocolumn,letterpaper]{article}

\usepackage{iccv}
\usepackage{times}
\usepackage{epsfig}
\usepackage{graphicx}
\usepackage{amsmath}
\usepackage{amssymb}
\usepackage{cite}
\usepackage{bm}
\usepackage{xspace}
\usepackage{microtype}

\usepackage{mathalfa}
\usepackage{graphicx}
\usepackage{graphbox}
\usepackage{array}
\usepackage{enumitem}
\usepackage{animate}
\usepackage{xspace} 
\usepackage{ctable}
\usepackage{multirow}
\usepackage{caption}

\newcommand{\update}[1]{{}{#1}}

\makeatletter
\renewcommand{\paragraph}{\@startsection{paragraph}{4}{\z@}{1.2ex plus   
0.5ex minus .2ex}{-1em}{\normalsize\bf}}
\makeatother

\newcommand{\mappingi}{\mathcal{T}_{\bm{i}}}
\newcommand{\mappingj}{\mathcal{T}_{\bm{j}}}
\newcommand{\xyzi}{\bm{x}_i}
\newcommand{\xyzj}{\bm{x}_j}
\newcommand{\xyzjhat}{\bm{\hat{x}}_j}
\newcommand{\uvw}{\bm{u}}

\newcommand{\mappingnet}{M_{\bm \theta}}
\newcommand{\nerf}{F_{\bm \theta}}
\newcommand{\latent}{\bm{\psi}}
\newcommand{\querypixel}{\bm{p}_i}
\newcommand{\predpixel}{\bm{\hat{p}}_j}
\newcommand{\pixelloc}{\bm{p}_i}
\newcommand{\predcolor}{\bm{\hat{C}}}
\newcommand{\gtcolor}{\bm{C}}
\newcommand{\lossflow}{\mathcal{L}_\text{flo}}
\newcommand{\losspho}{\mathcal{L}_\text{pho}}
\newcommand{\lossreg}{\mathcal{L}_\text{reg}}
\newcommand{\predflowij}{\bm{\hat{f}}_{i\rightarrow j}}
\newcommand{\gtflowij}{\bm{f}_{i\rightarrow j}}

\newcommand{\ray}{\bm r}
\newcommand{\origin}{\bm o}
\newcommand{\direction}{\bm d}

\usepackage[pagebackref=true,breaklinks=true,letterpaper=true,colorlinks,citecolor=blue,bookmarks=false]{hyperref}

\iccvfinalcopy %

\def\iccvPaperID{2206} %

\begin{document}

\title{Tracking Everything Everywhere All at Once}

\author{
Qianqian Wang$^{1,2}$ \quad
Yen-Yu Chang$^1$ \quad
Ruojin Cai$^1$ \quad
Zhengqi Li$^2$\\
Bharath Hariharan$^1$ \quad
Aleksander Holynski$^{2,3}$ \quad
Noah Snavely$^{1,2}$
\\[0.5em]
$^1$Cornell University \ \ \ 
$^2$Google Research \ \ \
$^3$UC Berkeley 
\vspace{-1em}
}

\twocolumn[{%
\renewcommand\twocolumn[1][]{#1}%
\maketitle

\begin{center}%
    \captionsetup{type=figure}%
    \includegraphics[width=\linewidth]{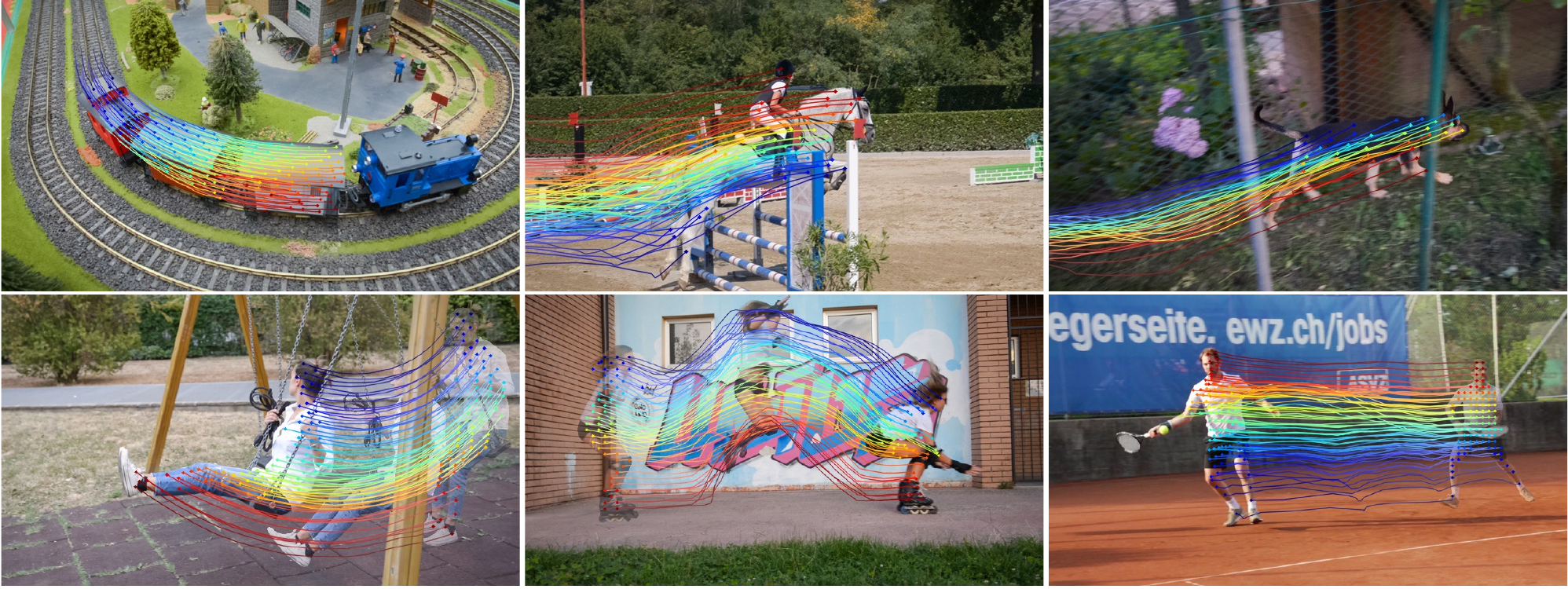}%
    \captionof{figure}{
    \small 
    We present a new method for estimating full-length motion trajectories for every pixel in every frame of a video, as illustrated by the motion paths shown above.
    For clarity, we only show sparse trajectories for foreground objects, though our method computes motion for \emph{all} pixels. Our method yields accurate, coherent long-range motion even for fast-moving objects, and robustly tracks through occlusions as shown in the \textit{dog} and \textit{swing} examples. For context, in the second row we depict the moving object at different moments in time.
    }%
    \label{fig:teaser}%
\end{center}%

}]
\maketitle

\begin{abstract}
    We present a new test-time optimization method for estimating dense and long-range motion from a video sequence. Prior optical flow or particle video tracking algorithms typically operate within limited temporal windows, struggling to track through occlusions and maintain global consistency of estimated motion trajectories. 
    We propose a complete and globally consistent motion representation, dubbed \emph{OmniMotion}, that allows for accurate, full-length motion estimation of every pixel in a video.
    OmniMotion represents a video using a quasi-3D canonical volume and performs pixel-wise tracking via bijections between local and canonical space. This representation allows us to ensure global consistency, track through occlusions, and model any combination of camera and object motion.  Extensive evaluations on the TAP-Vid benchmark and real-world footage show that our approach outperforms prior state-of-the-art methods by a large margin both quantitatively and qualitatively. See our project page for more results: \href{http://omnimotion.github.io/}{omnimotion.github.io}.
    
\end{abstract}

\section{Introduction}
\label{sec: intro}
Motion estimation methods have traditionally 
followed one of two dominant approaches:
sparse feature tracking and dense optical flow~\cite{sand2008particle}. While each type of method has proven effective for their respective applications, neither representation 
fully models the motion of a video: pairwise optical flow fails to capture motion trajectories over long temporal windows, 
and sparse tracking does not model the motion of all pixels. 

A number of approaches have sought to close this gap, i.e., to estimate both \emph{dense} and \emph{long-range} pixel trajectories in a video. These range from 
methods that simply chain together two-frame optical flow fields, to more recent approaches that directly predict per-pixel trajectories across multiple frames~\cite{harley2022particle}. 
Still, these methods all use limited context when estimating motion, 
disregarding information that is either temporally or spatially distant.
This locality can result in accumulated errors over long trajectories 
and spatio-temporal inconsistencies in the motion estimates.
Even when prior methods do consider long-range context~\cite{sand2008particle}, they operate in the 2D domain, resulting in a loss of tracking during occlusion events. 
All in all, producing both \emph{dense} and \emph{long-range} trajectories remains an open problem in the field, with three key challenges: (1) maintaining accurate tracks across long sequences, (2) tracking points through occlusions, and (3) maintaining
coherence in space and time.

In this work, we propose a holistic approach to video motion estimation that uses \emph{all} the information in a video to jointly estimate full-length motion trajectories for every pixel. 
Our method, which we dub \emph{OmniMotion}, uses a quasi-3D representation in which a canonical 3D volume is mapped to per-frame local volumes through a set of local-canonical bijections. These bijections serve as a flexible relaxation of dynamic multi-view geometry, modeling a combination of camera and scene motion.
Our representation guarantees cycle consistency, and can track all pixels, even while occluded (\emph{``Everything,~Everywhere"}). 
We optimize our representation per video to jointly solve for the motion of the entire video~\emph{``All~at~Once"}. Once optimized, our representation can be queried at any continuous coordinate in the video to receive a motion trajectory spanning the entire video. 

In summary, we propose an approach that: 1) produces globally consistent full-length motion trajectories for all points in an entire video,
2) can track points through occlusions, and 3) can tackle in-the-wild videos with any combination of camera and scene motion.
We demonstrate these strengths quantitatively on the TAP video tracking benchmark~\cite{doersch2022tap}, where we achieve state-of-the-art performance, outperforming all prior methods by a large margin.

\section{Related Work}
\paragraph{Sparse feature tracking.}
Tracking features~\cite{lowe2004distinctive,bay2008speeded} across images is essential for a wide range of applications such as Structure from Motion~(SfM)~\cite{schonberger2016structure, agarwal2011building,snavely2008modeling} and 
SLAM~\cite{durrant2006simultaneous}.  
While sparse feature tracking~\cite{lucas1981iterative,tomasi1991detection,shi1994good,detone2018superpoint} can establish long-range correspondence, this correspondence is limited to a set of distinctive interest points, and often restricted to rigid scenes. Hence, below we focus on work that can produce dense pixel motion for general videos.

\paragraph{Optical flow.}
Optical flow 
has traditionally been formulated as an optimization problem~\cite{horn1981determining,black1993framework,brox2009large,weinzaepfel2013deepflow}. However, recent advances 
have enabled direct prediction of optical flow using neural networks with improved quality and efficiency~\cite{fischer2015flownet,ilg2017flownet,sun2018pwc,hui2018liteflownet}. 
One 
leading method, RAFT~\cite{teed2020raft}, estimates flow through iterative updates of a flow field based on 4D correlation volumes. While optical flow methods allow for precise motion estimation between consecutive frames, they are not suited to long-range motion estimation: chaining pairwise optical flow into longer trajectories results in drift and fails to handle occlusions, while directly computing optical flow between distant frames (i.e., larger displacements) often results in temporal inconsistency~\cite{brox2010large,weinzaepfel2013deepflow}.
\update{Multi-frame flow estimation methods~\cite{ren2019fusion,janai2018unsupervised,irani1999multi,volz2011modeling} can address some limitations of two-frame optical flow, but still struggle to handle long-range motion.}

\paragraph{Feature matching.}
While optical flow methods are typically intended to operate on consecutive frames, other techniques can estimate dense correspondences between distant pairs of video frames~\cite{liu2010sift}.
Several methods learn such correspondences in a self- or weakly-supervised manner~\cite{caron2021emerging,li2019joint, wang2020learning,detone2018superpoint,vondrick2018tracking,rocco2018neighbourhood,bian2022learning,xu2021rethinking} using cues like cycle consistency~\cite{wang2019learning,zhou2016learning,jabri2020space}, while others~\cite{sun2021loftr,dusmanu2019d2,jiang2021cotr,truong2021learning, truong2020glu} use stronger supervision signals such as geometric correspondences generated from 3D reconstruction pipelines~\cite{li2018megadepth,schonberger2016structure}. 
However, pairwise matching approaches typically do not incorporate temporal context, which can lead to inconsistent tracking over long videos and poor occlusion handling.
\update{In contrast, our method produces smooth trajectories through occlusions.}

\update{
\paragraph{Pixel-level long-range tracking.}
A notable recent approach, PIPs~\cite{harley2022particle}, estimates multi-frame trajectories through occlusions by leveraging context within a small temporal window~(8 frames). 
However, to produce motion for videos longer than this temporal window, PIPs still must chain correspondences, a process that (1) is prone to drift and (2) will lose track of points that remain occluded beyond the 8-frame window.
Concurrent to our work, several works develop learning-based methods for predicting long-range pixel-level tracks in a feedforward manner. MFT~\cite{neoral2023mft} learns to select the most reliable sequences of optical flows to perform long-range tracking. TAPIR~\cite{doersch2023tapir} tracks points by employing a matching stage inspired by TAP-Net~\cite{doersch2022tap} and a refinement stage inspired by PIPs~\cite{harley2022particle}. CoTracker~\cite{karaev2023cotracker} proposes a flexible and powerful tracking algorithm with a transformer architecture to track points throughout a video. 
Our contribution is complementary to these works: the output of any of these methods can be used as the input supervision when optimizing a global motion representation.
}

\paragraph{Video-based motion optimization.}
Most conceptually related to our approach are classical methods that optimize motion globally over an entire video~\cite{lezama2011track,sand2008particle,rubinstein2012towards,sundaram2010dense,chang2013video,badrinarayanan2010label,wang1993layered,sun2010layered}. 
Particle video, for instance, produces a set of semi-dense long-range trajectories (called \emph{particles}) from initial optical flow fields~\cite{sand2008particle}.
However, it does not 
track through occlusions; an occluded entity will be treated as a different particle when it re-appears. Rubinstein et al.~\cite{rubinstein2012towards} further proposed a combinatorial assignment approach that can track through occlusion and generate longer trajectories.
However, 
this method only produces semi-dense tracks for videos with simple motion, whereas our method estimates long-range motion for all pixels in general videos. 
Also related is ParticleSfM~\cite{zhao2022particlesfm}, which optimizes long-range correspondences from pairwise optical flows. Unlike our approach, ParticleSfM focuses on camera pose estimation within an SfM framework, where only correspondences from static regions are optimized, and dynamic objects are treated as outliers.

\paragraph{Neural video representations.} 
While our method shares similarities with recent methods that model videos using coordinate-based multi-layer perceptrons~(MLPs)~\cite{mildenhall2021nerf,sitzmann2020implicit,tancik2020fourier},
prior work has primarily focused on problems such as novel view synthesis~\cite{park2021nerfies,li2022dynibar,li2021neural,park2021hypernerf,xian2021space} and video decomposition~\cite{kasten2021layered,ye2022deformable}. 
In contrast, our work 
targets the challenge of dense, long-range motion estimation. 
Though some methods for dynamic novel view synthesis produce 2D motion as a by-product, these systems require known camera poses and the resulting motion is often erroneous~\cite{gao2022monocular}.
\update{Some dynamic reconstruction methods~\cite{yang2021lasr, yang2021viser,wu2022casa,cai2022neural} can also produce 2D motion, but these are often object-centric with a focus on articulated objects.}
Alternatively, video decomposition-based representations such as Layered Neural Atlases~\cite{kasten2021layered} and Deformable Sprites~\cite{ye2022deformable} 
solve for a mapping between each frame and a global texture atlas. Frame-to-frame correspondence can be derived by inverting this mapping, but this process is expensive and unreliable. Furthermore, these methods are limited to representing videos using a limited number of layers with fixed ordering, restricting their ability to model complex, real-world videos.

\section{Overview}
We propose a test-time optimization approach for estimating dense and long-range motion from a video sequence. Our method takes as input a collection of frames and pairwise noisy motion estimates (e.g., optical flow fields), and uses these to solve for a complete, globally consistent motion representation for the entire video. Once optimized, our representation can be queried with any pixel in any frame to produce a smooth, accurate motion trajectory across the full video. Our method identifies when points are occluded, and even tracks points through occlusions.
In the following sections, we describe our underlying representation, dubbed \emph{OmniMotion}~(Sec.~\ref{sec: omnimotion}), then describe our optimization process~(Sec.~\ref{sec:optimization}) for recovering this representation from a video.

\section{OmniMotion representation}
\label{sec: omnimotion}

\begin{figure*}
\begin{center}%
    \captionsetup{type=figure}%
    \includegraphics[width=\linewidth]{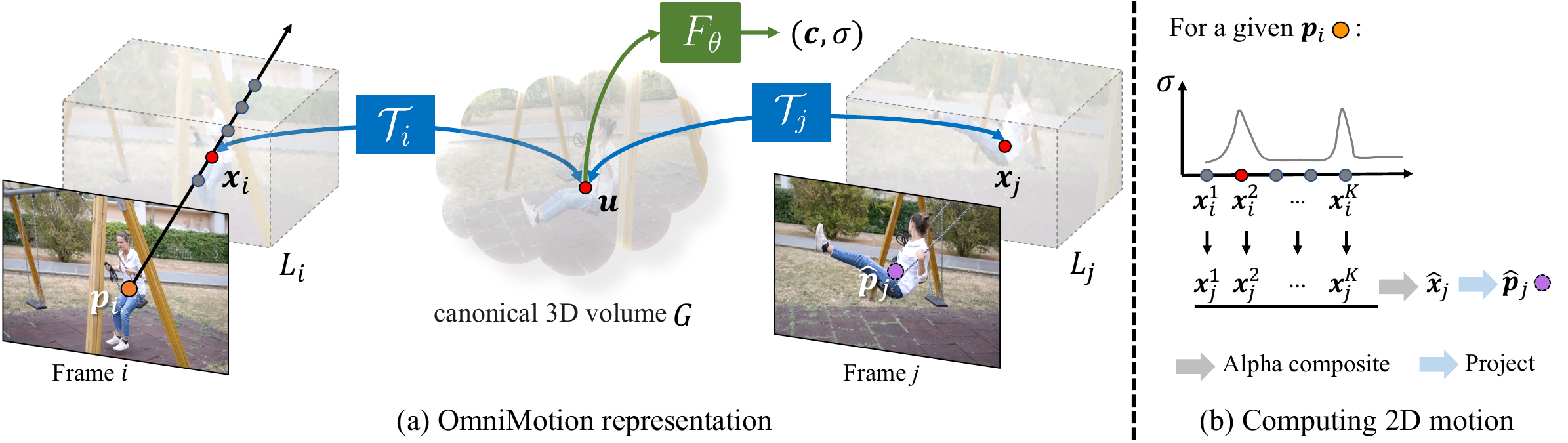}%
    \captionof{figure}{
    \small 
    \textit{Method overview}. (a) Our OmniMotion representation is comprised of a canonical 3D volume $G$ and a set of bijections $\mappingi$ that map between each frame's local volume $L_i$ and the canonical volume $G$. 
    Any local 3D location $\xyzi$ in frame $i$ can be mapped to its corresponding canonical location $\uvw$ through $\mappingi$, and then mapped back to another frame $j$ as $\xyzj$ through the inverse mapping $\mappingj^{-1}$. 
    Each location $\uvw$ in $G$ 
    is associated with a color $\bm c$ and density $\sigma$, computed using a coordinate-based MLP $\nerf$. (b) To compute the corresponding 2D location for a given query point $\querypixel$ mapped from frame $i$ to $j$, we shoot a ray into $L_i$ and sample a set of points $\{\xyzi^k\}_{k=1}^K$, which are then mapped first to the canonical space to obtain their densities, and then to frame $j$ to compute their corresponding local 3D locations $\{\xyzj^k\}_{k=1}^K$. These points $\{\xyzj^k\}_{k=1}^K$ are then alpha-composited and projected to obtain the 2D corresponding location $\predpixel$.
    }%
    \label{fig:overview}%
\end{center}%
\end{figure*}

As discussed in Sec.~\ref{sec: intro}, classical motion representations, such as pairwise optical flow, lose track of objects when they are occluded, and can produce inconsistencies when correspondences are composed over multiple frames.
To obtain accurate, consistent tracks even through occlusion, we therefore need a \emph{global} motion representation, i.e., a data structure that 
encodes the trajectories of all points in a scene.
One such global representation 
is a decomposition of a scene into a set of discrete, depth-separated layers~\cite{kasten2021layered,ye2022deformable}. However, most real-world 
scenes cannot be represented as a set of fixed, ordered layers: e.g., consider the simple case of an object rotating in 3D.
At the other extreme is full 3D reconstruction that disentangles 3D scene geometry, camera pose and scene motion. 
This, however, is an extremely ill-posed problem.
Thus, we ask: can we accurately track real-world motion without explicit dynamic 3D reconstruction?

We answer this question using our proposed representation, \emph{OmniMotion}~(illustrated in Fig.~\ref{fig:overview}).
OmniMotion represents the scene in a video as a canonical 3D volume that is mapped to local volumes for each frame through local-canonical bijections.
The local-canonical bijections are parametrized as neural networks and capture both camera and scene motion without disentangling the two.
As such, the video can be considered as a rendering of the resulting local volumes from a fixed, static camera.

Because OmniMotion does not explicitly disentangle camera and scene motion, the resulting representation is not a physically accurate 3D scene reconstruction.
Instead we call it a \emph{quasi}-3D representation.
This relaxation of dynamic 
multi-view geometry allows us to sidestep ambiguities that make dynamic 3D reconstruction challenging.
Yet we retain properties needed for  consistent and accurate long-term tracking through occlusion:
first, by establishing  bijections between each local frame and a canonical frame, OmniMotion guarantees globally cycle-consistent 3D mappings across all local frames, which
emulates the one-to-one correspondences between real-world, metric 3D reference frames.
Second, OmniMotion retains information about all scene points that are projected onto each pixel, along with their relative depth ordering, enabling points to be tracked even when they are temporarily occluded from view.

In the following sections, we 
describe our quasi-3D canonical volume and 3D bijections, and then describe how they can be used to compute motion between any two frames.

\subsection{Canonical 3D volume}

We represent a video's content using a canonical volume $G$ that acts %
as a three-dimensional atlas of the observed scene.
As in NeRF~\cite{mildenhall2021nerf}, 
we define a coordinate-based network $\nerf$ over $G$ that maps each canonical 3D coordinate $\uvw\in G$ to a density $\sigma$ and color $\bm c$. 
The density stored in $G$ is key, as it tells us where the surfaces are in canonical space.
Together with the 3D bijections, this allows us to track surfaces over multiple frames as well as reason about occlusion relationships.
The color stored in $G$ allows us to compute a photometric loss during optimization.

\subsection{3D bijections}
We define a continuous bijective mapping $\mappingi$ that 
maps 3D points $\xyzi$ from each local coordinate frame $L_i$ to the canonical 3D coordinate frame as $\uvw = \mappingi(\xyzi)$, where $i$ is the frame index. Note that the canonical coordinate $\uvw$ is time-independent and can be viewed as a globally consistent ``index'' for a particular scene point or 3D trajectory across time. %
By composing these bijective mappings and their inverses, we can map a 3D point from one local 3D coordinate frame ($L_i$) to another  ($L_j$):
\begin{equation}
    \xyzj = \mappingj^{-1} \circ \mappingi(\xyzi).
    \label{eq:local-local mapping}
\end{equation}
Bijective mappings ensure that the resulting correspondences between 3D points in individual frames are all cycle consistent, as they arise from the same canonical point.

To allow for expressive maps that can capture real-world motion, we parameterize these bijections as invertible neural networks~(INNs).
Inspired by recent work on homeomorphic shape modeling~\cite{paschalidou2021neural, lei2022cadex}, we use Real-NVP~\cite{dinh2016density} due to its simple formulation and analytic invertibility.
Real-NVP builds bijective mappings by composing 
simple bijective transformations called affine coupling layers.
An affine coupling layer splits the input into two parts; the first part is left unchanged, but is used to parametrize an affine transformation that is applied to the second part.

We modify this architecture to also condition on a per-frame latent code $\latent_i$~\cite{paschalidou2021neural, lei2022cadex}. 
Then all invertible mappings $\mappingi$ are parameterized by the same invertible network $\mappingnet$, but with different latent codes:
$\mappingi(\cdot) = \mappingnet(\cdot; \latent_i)$.

\subsection{Computing frame-to-frame motion}

Given this representation, we now describe how we can compute 2D motion for any query pixel $\querypixel$ in frame $i$. 
Intuitively, we ``lift'' the query pixel to 3D by sampling points on a ray, ``map'' these 3D points to a target frame $j$ using bijections $\mappingi$ and $\mappingj$, ``render'' these mapped 3D points from the different samples through alpha compositing, and finally ``project'' back to 2D to obtain a putative correspondence.

Specifically, 
since we assume that camera motion is subsumed by the local-canonical bijections $\mappingi$, we simply use a fixed, orthographic camera.
The ray at $\querypixel$ can then be defined as $\ray_i(z) = \origin_i+ z\direction$, where $\origin_i = [\querypixel, 0]$ and $\direction = [0, 0, 1]$. We sample $K$ samples on the ray $\{\bm{x}_i^k\}$, which are equivalent to appending a set of depth values $\{z_i^k\}_{k=1}^K$ to $\pixelloc$. 
Despite not being a physical camera ray, it captures the notion of multiple surfaces at each pixel and suffices to handle occlusion.

Next we obtain densities and colors for these samples by mapping them to the canonical space and then querying the density network $\nerf$.
Taking the $k$-th sample $\xyzi^k$ as an example, its density and color can be written as 
$(\sigma_k, \bm c_k) = \nerf(\mappingnet(\xyzi^k; \latent_i))$. 
We can also map each sample along the ray to a corresponding 3D location $\xyzj^k$ in frame $j$ (Eq.~\ref{eq:local-local mapping}).

We can now aggregate the correspondences $\xyzj^k$ from all samples to produce a single correspondence $\xyzjhat$.
This aggregation is similar to how the colors of sample points are aggregated in NeRF: we use alpha compositing, with the alpha value for the $k$-th sample as
$\alpha_k = 1 - \exp(-\sigma_k)$. 
We then compute $\xyzjhat$ as:
\begin{align}
    \xyzjhat = \sum_{k=1}^K T_k \alpha_k\xyzj^k, 
    \text{ where }
     T_k = \prod_{l=1}^{k-1}(1 - \alpha_l)
    \label{eq:alpha blending}
\end{align}
A similar process is used to composite $\bm c^k$ to get the image-space color $\predcolor_i$ for $\querypixel$.
$\xyzjhat$ is then projected using
our stationary orthographic camera model to yield the predicted 2D corresponding location $\predpixel$ for the query location $\querypixel$.

\section{Optimization}
\label{sec:optimization}
Our optimization process takes as input a video sequence and a collection of noisy correspondence predictions (from an existing method) as guidance, and generates a complete, globally consistent motion estimate for the entire video.

\subsection{Collecting input motion data} 
For most of our experiments, we use RAFT~\cite{teed2020raft} 
to compute input pairwise correspondence. We also experimented with another dense correspondence method, TAP-Net~\cite{doersch2022tap}, and demonstrate in our evaluation that our approach consistently works well given different types of input correspondence.
Taking RAFT as an example, we begin by exhaustively computing all pairwise optical flows. Since optical flow methods can produce significant errors under large displacements, we apply cycle consistency and appearance consistency checks to filter out spurious correspondences. We also optionally augment the flows through chaining, when deemed reliable.
Additional details about our flow collection process 
are provided in the supplemental material.
Despite the filtering, the (now incomplete) flow fields 
remain noisy and inconsistent. 
We now introduce our optimization method that consolidates these noisy, incomplete pairwise motion into complete and accurate long-range motion. %

\subsection{Loss functions}
\label{sec: loss_functions}
Our primary loss function is a flow loss. We minimize the mean absolute error~(MAE) between the predicted flow $\predflowij = \predpixel - \querypixel$ from our optimized representation and the supervising input flow $\gtflowij$ derived from running optical flow:
\begin{equation}
    \lossflow = \sum_{\gtflowij \in \Omega_f} ||\predflowij - \gtflowij||_1
\end{equation}
where $\Omega_f$ is the set of all the filtered pairwise flows.
In addition, we minimize a photometric loss defined as the mean squared error~(MSE) between the predicted color $\predcolor_i$ and the observed color $\gtcolor_i$ in the source video frame:
\begin{equation}
    \losspho = \sum_{(i, \bm p)\in \Omega_p}||\predcolor_i(\bm{p}) - \gtcolor_i(\bm{p})||_2^2
\end{equation}
where $\Omega_p$ is the set of all pixel locations over all frames.
Last, to ensure temporal smoothness of the 3D motion estimated by $\mappingnet$, we apply a regularization term that penalizes large accelerations. 
Given a sampled 3D location $\bm{x}_i$ in frame $i$, we map it to frame $i-1$ and frame $i+1$ using Eq.~\ref{eq:local-local mapping}, yielding 3D points $\bm{x}_{i-1}$ and $\bm{x}_{i+1}$ respectively, and then minimize 3D acceleration as in~\cite{li2021neural}:
\begin{equation}
    \lossreg = \sum_{(i, \bm x)\in\Omega_x}||\bm x_{i+1} + \bm x_{i-1} - 2\bm x_i||_1
\end{equation}
where $\Omega_x$ is the union of local 3D spaces for all frames.
Our final combined loss can be written as:
\begin{equation}
    \mathcal{L} = \lossflow + \lambda_{\text{pho}}\losspho + \lambda_{\text{reg}}\lossreg
\end{equation}
where weights $\lambda$ control the relative importance of each term.

The intuition behind this optimization is to leverage the bijections to a single canonical volume $G$,
photo consistency, and the natural spatiotemporal smoothness provided by the coordinate-based networks $\mappingnet$ and $\nerf$  to reconcile inconsistent pairwise flow and fill in missing content in the correspondence graphs.

\subsection{Balancing supervision via hard mining}
The exhaustive pairwise flow input maximizes the useful motion information available to the optimization stage. 
However, this approach, especially when coupled with the flow-filtering process,
can result in an unbalanced collection of motion samples in dynamic regions. 
Rigid background regions typically have many reliable pairwise correspondences, while fast-moving and deforming foreground objects often have many fewer reliable correspondences after filtering, especially between distant frames. 
This imbalance can lead the network to focus entirely on dominant (simple) background motions, and ignore the challenging moving objects that represent a small portion of the supervisory signal. 

To address this issue, we propose a simple 
strategy for mining hard examples during training. 
Specifically, we periodically cache flow predictions 
and compute error maps by calculating the Euclidean distance between the predicted and input flows. 
During optimization, we guide sampling such that 
regions with high errors are sampled more frequently. 
We compute these error maps on consecutive frames, where we assume our supervisory optical flow is most reliable.
Please refer to the supplement for more details.

\subsection{Implementation details}
\paragraph{Network.} Our mapping network $\mappingnet$ consists of six affine coupling layers. We apply positional encoding~\cite{mildenhall2021nerf,tancik2020fourier} with 4 frequencies to
each layer's input coordinates before computing the scale and translation.
We use a single 2-layer MLP with 256 channels implemented as a GaborNet~\cite{fathony2021multiplicative} 
to compute the latent code $\latent_i$ for each frame $i$. The input to this MLP is the time $t_i$.
The dimensionality of latent code $\latent_i$ is 128. The canonical representation $\nerf$ is also implemented as a GaborNet, but with 3 layers of 512 channels.

\paragraph{Representation.} We normalize all pixel coordinates $\pixelloc$ to the range $[-1, 1]$, and set the near and far depth range 
to $0$ and $2$, defining a local 3D space for each frame as $[-1, 1]^2 \times [0, 2]$. While our method 
can place content at arbitrary locations in the canonical space $G$, we initialize mapped canonical locations given by $\mappingnet$ to be roughly within a unit sphere to ensure well-conditioned input to the density/color network $\nerf$. 
To improve numerical stability during training, we apply the contraction operation in mip-NeRF 360~\cite{barron2022mip} to canonical 3D coordinates $\uvw$ before passing them to $\nerf$.

\paragraph{Training.} We train our representation on each video sequence with Adam~\cite{kingma2014adam} for 200K iterations.
In each training batch, we sample 256 
correspondences from 8 pairs of images, for a total of 1024 correspondences. We sample $K=32$ points on each ray using stratified sampling. More training details can be found in the supplemental material.

\section{Evaluation}
\subsection{Benchmarks}
We evaluate our method on the TAP-Vid benchmark~\cite{doersch2022tap}, which is designed to evaluate the performance of point tracking across long video clips. 
TAP-Vid consists of both real-world videos with accurate human annotations of point tracks and synthetic videos with perfect ground-truth point tracks. 
Each point track is annotated through the entire video, and is labeled as occluded when not visible. 

\paragraph{Datasets.} We evaluate on the following datasets from TAP-Vid: 
1) \textbf{DAVIS}, a real dataset of 30 videos from the DAVIS 2017 validation set~\cite{pont20172017}, with clips ranging from 34-104 frames and an average of 21.7 point track annotations per video. 
2) \textbf{Kinetics}, a real dataset of 1,189 videos each with 250 frames
from the Kinetics-700-2020 validation set~\cite{carreira2017quo} with an average of 26.3 point track annotations per video. 
To make evaluation tractable for test-time optimization approaches like ours, 
we randomly sample a subset of 100 videos and evaluate all methods on this subset. 3) \textbf{RGB-Stacking}~\cite{lee2021beyond}, a synthetic dataset of 50 videos each with 250 frames 
and 30 tracks.
We exclude the synthetic Kubric dataset~\cite{greff2022kubric} as it is primarily intended for training. 
For quantitative evaluation, we adhere to the TAP benchmark protocol and evaluate all methods at 256$\times$256 resolution, but all qualitative results are run at a higher resolution (480p).

\paragraph{Evaluation metrics.} 
Following the TAP-Vid benchmark, we report both the position 
and occlusion accuracy of predicted tracks. We also introduce a new metric measuring temporal coherence. Our evaluation metrics include:
\vspace{-0.3em}
\begin{itemize}[leftmargin=*]
  \setlength\itemsep{-0.05em}
  \item \boldsymbol{$<\delta^x_\textbf{avg}$} measures the average position accuracy of visible points across 5 thresholds: 1, 2, 4, 8, and 16 pixels. The accuracy $<\delta^x$ at each threshold $\delta^x$ is the fraction of points that are within $\delta^x$ pixels of their ground truth position.
  \item \textbf{Average Jaccard (AJ)} evaluates both occlusion and position accuracy on the same thresholds as $<\delta^x_\textrm{avg}$. It categorizes predicted point locations as true positives, false positives, and false negatives, and is defined as the ratio of true positives to all points. True positives are points within the threshold of visible ground truth points. False positives are points that are predicted as visible, but where the ground truth is occluded or beyond the threshold. False negatives are ground truth visible points that are predicted as occluded or are beyond the threshold.
  \item \textbf{Occlusion Accuracy (OA)} evaluates the accuracy of the visibility/occlusion prediction at each frame.
  \item \textbf{Temporal Coherence (TC)} evaluates the temporal coherence of the tracks by measuring the $L_2$ distance between the acceleration of groundtruth tracks and predicted tracks. The acceleration is measured as the flow difference between two adjacent frames for visible points.
\end{itemize}

\subsection{Baselines}
We compare OmniMotion to various types of dense correspondence methods, including optical flow, feature matching, and multi-frame trajectory estimation as follows: 

\vspace{0.5em}\noindent\textbf{RAFT}~\cite{teed2020raft} is a state-of-the-art two-frame flow method. 
We consider two ways to use RAFT to generate multi-frame trajectories at test time: 
1) chaining RAFT predictions between consecutive frames into longer tracks, which we call \textbf{RAFT-C}, and 
2) directly computing RAFT flow between any (non-adjacent) query and target frames (\textbf{RAFT-D}). 
When generating trajectories using RAFT-D, we always use the previous flow prediction as initialization for the current frame.%

\vspace{0.5em}\noindent\textbf{PIPs}~\cite{harley2022particle} is a method for estimating multi-frame point trajectories that can handle occlusions. By default, the method uses a temporal window of 8 frames, and longer trajectories must be generated through chaining. 
We used the official implementation of PIPs to perform chaining.

\vspace{0.5em}\noindent\textbf{Flow-Walk}~\cite{bian2022learning} uses a multi-scale contrastive random walk to learn space-time correspondences by encouraging cycle consistency across time. 
Similar to RAFT, we report both chained and direct correspondence computation as \textbf{Flow-Walk-C} and \textbf{Flow-Walk-D}, respectively.

\vspace{0.5em}\noindent\textbf{TAP-Net}~\cite{doersch2022tap}  uses a cost volume to predict the location of a query point in a single target frame, along with a scalar occlusion logit. %

\vspace{0.5em}\noindent\textbf{Deformable Sprites}~\cite{ye2022deformable} is a layer-based video decomposition method.
Like our method, it uses a per-video
test-time optimization. However, 
it does not directly produce frame-to-frame correspondence, as the mapping from each frame to a canonical texture image is non-invertible. 
A nearest neighbor search in texture image space is required to find correspondence. 
Layered Neural Atlases~\cite{kasten2021layered} shares similarities to Deformable Sprites, but requires semantic segmentation masks as input, so we opt to compare to Deformable Sprites.

PIPs, TAP-Net and Deformable Sprites directly predict occlusion, but 
RAFT and Flow-Walk do not.
Therefore we follow prior work~\cite{doersch2022tap,xu2021rethinking} and use a cycle consistency check with a threshold of $48$ pixels to produce occlusion predictions for these methods. 
For our method, we detect occlusion by first mapping the query point to its corresponding 3D location in the target frame, then checking the transmittance of that 3D location in the target frame. 

\begin{figure*}[t!]
    \centering
    \includegraphics[width=\linewidth]{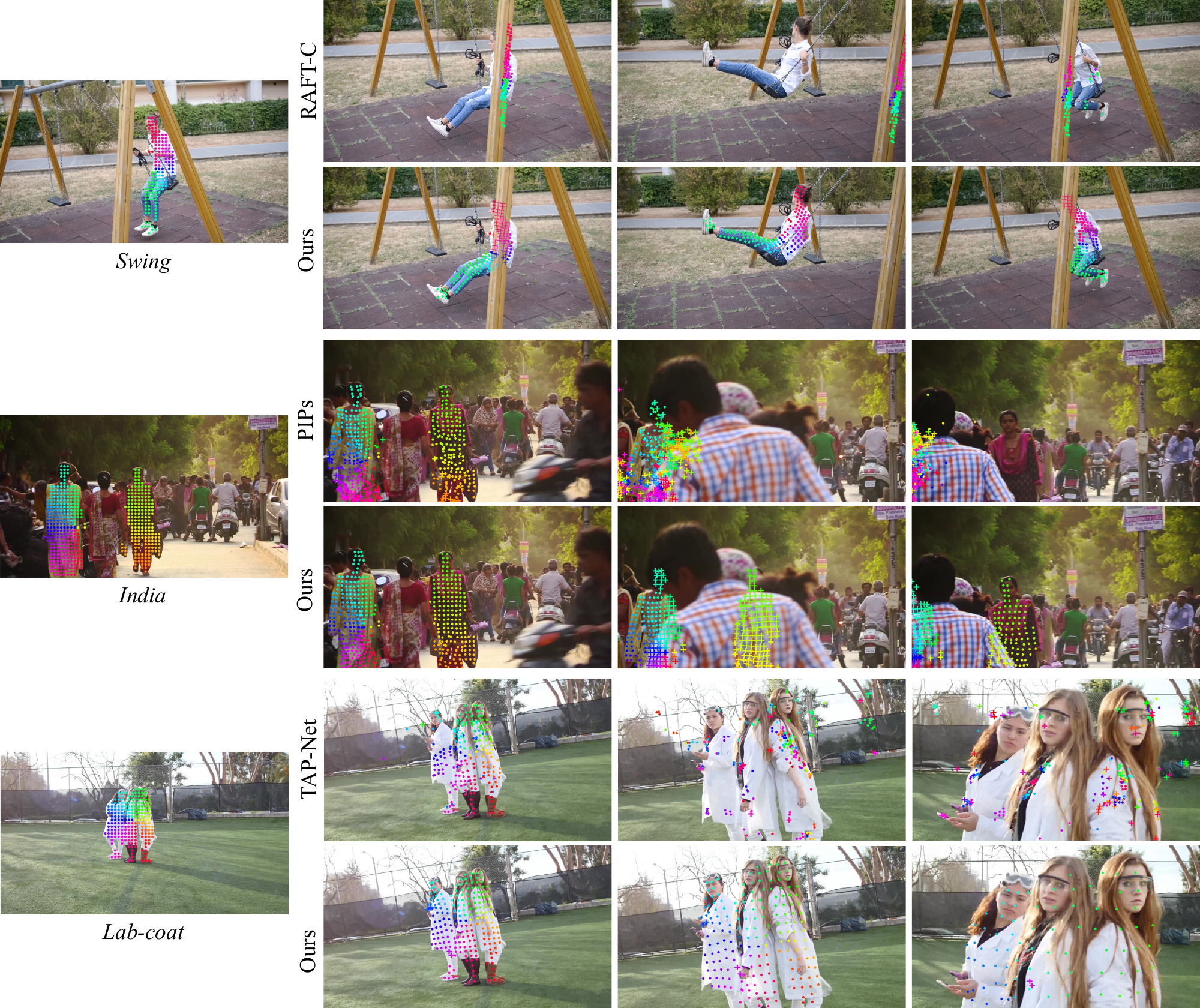}
    \caption{\small \emph{Qualitative comparison of our method and baselines on DAVIS}~\cite{pont20172017}. The leftmost image shows query points in the first frame, and the right three images show tracking results over time. 
    Notably, our method tracks successfully through the occlusion events in \textit{swing} and \textit{india}, while baseline methods fail. Our method additionally detects occlusion~(marked as cross ``$+$'') and gives plausible location estimates even when a point is occluded. Please refer to the supplemental video for better visualizations of tracking accuracy and coherence.
    }
    \label{fig: qualitative}
\end{figure*}

\subsection{Comparisons}
\begin{table*}[ht]
{\resizebox{\linewidth}{!}{
\small
\begin{tabular}{lcccccccccccc}
\toprule
\multirow{2}{*}{\textbf{Method}} & 
\multicolumn{4}{c}{\textbf{Kinetics}} & \multicolumn{4}{c}{\textbf{DAVIS}} & \multicolumn{4}{c}{\textbf{RGB-Stacking}} \\ 
\cmidrule(lr){2-5} \cmidrule(lr){6-9} \cmidrule(lr){10-13} 
    & AJ~$\uparrow$ & $<\delta^x_\textrm{avg}$~$\uparrow$ & OA~$\uparrow$ & TC~$\downarrow$ & AJ~$\uparrow$ & $<\delta^x_\textrm{avg}$~$\uparrow$ & OA~$\uparrow$ & TC~$\downarrow$ & AJ~$\uparrow$ & $<\delta^x_\textrm{avg}$~$\uparrow$ & OA~$\uparrow$ & TC~$\downarrow$ \\ 
\midrule
RAFT-C~\cite{teed2020raft} & $31.7$ & $51.7$ & $84.3$ & $0.82$ & $30.7$ & $46.6$ & $80.2$ & $0.93$ & $42.0$ & $56.4$ & $91.5$ & $0.18$  \\
RAFT-D~\cite{teed2020raft} & $50.6$ & $66.9$ & $85.5$ & $3.00$ & $34.1$ & $48.9$ & $76.1$ & $9.83$ & $72.1$ & $\underline{85.1}$ & $92.1$ & $1.04$ \\
TAP-Net~\cite{doersch2022tap} & $48.5$ & $61.7$ & $86.6$ & $6.65$ & $38.4$ & $53.4$ & $81.4$ & $10.82$ & $61.3$ & $73.7$ & $91.5$ & $1.52$ \\
PIPs~\cite{harley2022particle} & $39.1$ & $55.3$ & $82.9$ & $1.30$ & $39.9$ & $56.0$ & $81.3$ & $1.78$ & $37.3$ & $50.6$ & $89.7$ & $0.84$ \\
Flow-Walk-C~\cite{bian2022learning} & $40.9$ & $55.5$ & $84.5$ & $\underline{0.77}$ & $35.2$ & $51.4$ & $80.6$ & $0.90$ & $41.3$ & $55.7$ & $\underline{92.2}$ & $\underline{0.13}$ \\
Flow-Walk-D~\cite{bian2022learning} & $46.9$ & $65.9$ & $81.8$ & $3.04$ & $24.4$ & $40.9$ & $76.5$ & $10.41$ & $66.3$ & $82.7$ & $91.2$ & $0.47$ \\
Deformable-Sprites~\cite{ye2022deformable} & $25.6$ & $39.5$ & $71.4$ & $1.70$ & $20.6$ & $32.9$ & $69.7$ & $2.07$ & $45.0$ & $58.3$ & $84.0$ & $0.99$ \\
\midrule
Ours~(TAP-Net) & $\underline{53.8}$ & $\underline{68.3}$ & $\underline{88.8}$ & $\underline{0.77}$ & $\underline{50.9}$ & $\underline{66.7}$ & $\mathbf{85.7}$ & $\underline{0.86}$ & $\underline{73.4}$ & $84.1$ & $\underline{92.2}$ & $\mathbf{0.11}$ \\
Ours~(RAFT) & $\mathbf{55.1}$ & $\mathbf{69.6}$ & $\mathbf{89.6}$ & $\mathbf{0.76}$ & $\mathbf{51.7}$ & $\mathbf{67.5}$ & $\underline{85.3}$ & $\mathbf{0.74}$ & $\mathbf{77.5}$ & $\mathbf{87.0}$ & $\mathbf{93.5}$ & $\underline{0.13}$ \\
\bottomrule
\end{tabular}
}}%
\caption{\small \emph{Quantitative comparison of our method and baselines on the TAP-Vid benchmark}~\cite{doersch2022tap}. We refer to our method as \emph{Ours}, and present two variants, \emph{Ours (TAP-Net)} and \emph{Ours (RAFT)}, which are optimized using input pairwise correspondences from TAP-Net~\cite{doersch2022tap} and RAFT~\cite{teed2020raft}, respectively.  Both \emph{Ours} and \emph{Deformable Sprites}~\cite{ye2022deformable} estimate global motion via test-time optimization on each individual video, while all other methods estimate motion locally in a feed-forward manner.
Our method notably improves the quality of the input correspondences, achieving the best position accuracy, occlusion accuracy, and temporal coherence among all methods tested.
}
\vspace{-0.1in}
\label{tab: quantitive}
\end{table*}

\paragraph{Quantitative comparisons.} We compare our method to baselines on the TAP-Vid benchmark in Tab.~\ref{tab: quantitive}. Our method achieves the best position accuracy, occlusion accuracy, and temporal coherence consistently across different datasets. 
Our method works well with different input pairwise correspondences from RAFT and TAP-Net, and provides consistent improvements over both of these base methods.

Compared to approaches that directly operate on (non-adjacent) pairs of query and target frames like RAFT-D, TAP-Net, and Flow-Walk-D, our method achieves significantly better temporal coherence due to our globally consistent representation.
Compared to flow chaining methods like RAFT-C, PIPs, and Flow-Walk-C, our method has better tracking performance, especially on longer videos. 
Chaining methods accumulate errors over time and are not robust to occlusion. 
Although PIPs considers a wider temporal window (8 frames) for better occlusion handling, it fails to track a point if it stays occluded beyond the entire temporal window. 
In contrast, OmniMotion can track points through extended occlusions. 
Our method also outperforms the test-time optimization approach Deformable Sprites~\cite{ye2022deformable}. 
Deformable Sprites decomposes a video using a predefined two or three layers with fixed ordering and models the background as a 2D atlas with a per-frame homography, limiting its ability to fit to videos with complex camera and scene motion.

\paragraph{Qualitative comparisons.}
We compare our method qualitatively to our baselines in Fig.~\ref{fig: qualitative}. We highlight our ability to identify and track through (long) occlusion events while also providing plausible locations for points during occlusion, as well as handling large camera motion parallax. 
Please refer to the supplementary video for animated comparisons.

\subsection{Ablations and analysis}
\begin{table}[t]
\centering

{\resizebox{0.85\linewidth}{!}{
\small 
\begin{tabular}{lcccc}
\toprule

\textbf{Method} & AJ~$\uparrow$ & $<\delta^x_\textrm{avg}$~$\uparrow$ & OA~$\uparrow$ & TC~$\downarrow$  \\ 
\midrule
No invertible & $12.5$ & $21.4$ & $76.5$ & $0.97$\\
No photometric & $42.3$ & $58.3$ & $84.1$ & $0.83$\\
Uniform sampling & $47.8$ & $61.8$ & $83.6$ & $0.88$ \\
\midrule
Full & $\mathbf{51.7}$ & $\mathbf{67.5}$ & $\mathbf{85.3}$ & $\mathbf{0.74}$ \\
\bottomrule
\end{tabular}
}}%

\caption{
\small
Ablation study on DAVIS~\cite{pont20172017}.
}
\vspace{-0.1in}
\label{tab: ablation}
\end{table}
\paragraph{Ablations.}
We perform ablations to verify the effectiveness of our design decisions in Tab.~\ref{tab: ablation}. \emph{No invertible} is a model variant that replaces our invertible mapping network $\mappingnet$ with separate forward and backward mapping networks between local and canonical frames (i.e., without the strict cycle consistency guarantees of our proposed bijections). 
While we additionally add a cycle consistency loss for this ablation, it still fails to construct a meaningful canonical space, and can only represent simple motions of the static background. 
\emph{No photometric} is a version that omits the photometric loss $\losspho$; the reduced performance suggests the importance of photoconsistency for refining motion estimates. 
\emph{Uniform sampling} replaces our hard-mining sampling strategy with a uniform sampling strategy, which we found leads to an inability to capture fast motion.%

\paragraph{Analysis.}
\begin{figure}[t!]
    \centering
    \includegraphics[width=\linewidth]{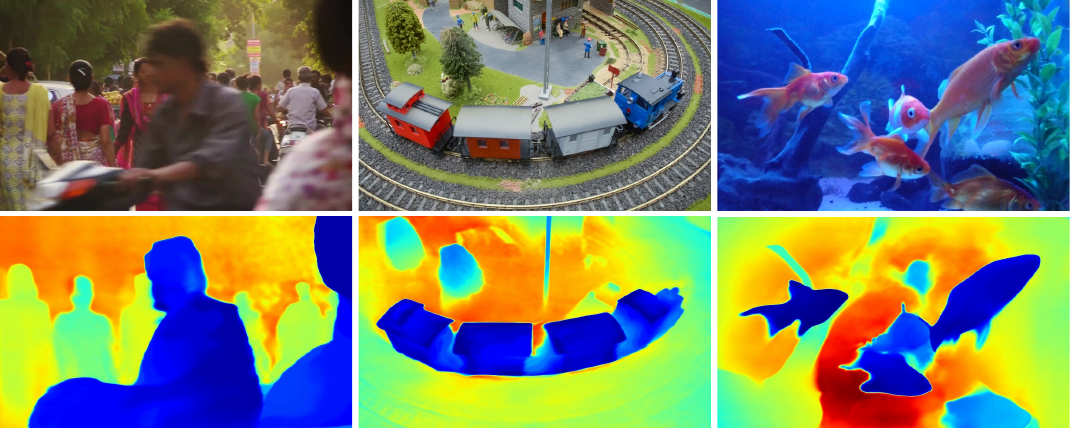}\\
    \caption{\small Pseudo-depth maps extracted from our representation, where blue indicates closer objects and red indicates further.}
    \label{fig: pseudo depth}
\end{figure}
In Fig.~\ref{fig: pseudo depth}, we show pseudo-depth maps generated from our model to demonstrate the learned depth ordering. 
Note that these maps do not correspond to physical depth, nonetheless, they demonstrate that using only photometric and flow signals, our method is able to sort out the relative ordering between different surfaces, which is crucial for tracking through occlusions. More ablations and analyses are provided in the supplemental material.

\section{Limitations}
Like many motion estimation methods, our method struggles with rapid and highly non-rigid motion as well as thin structures. In these scenarios, pairwise correspondence methods can fail to provide enough reliable correspondences for our method to compute accurate global motion. %

\update{
In addition, due to the highly non-convex nature of the underlying optimization problem, we observe that our optimization process can be sensitive to initialization for certain difficult videos. This can result in sub-optimal local minima, e.g., incorrect surface ordering or duplicated objects in canonical space that can sometimes be hard to correct through optimization. 
}

Finally, our method in its current form can be computationally expensive. 
First, the flow collection process involves computing all pairwise flows exhaustively, which scales quadratically with respect to the sequence length. 
However, we believe the scalability of this process can be improved by exploring more efficient alternatives to exhaustive matching, e.g., vocabulary trees or keyframe-based matching, drawing inspiration from the Structure from Motion and SLAM literature. 
Second, like other methods that utilize neural implicit representations~\cite{mildenhall2021nerf}, our method involves a relatively long optimization process. 
\update{Recent research in this area~\cite{mueller2022instant,tancik2022block} may help accelerate this process and allow further scaling to longer sequences.}

\section{Conclusion}
We proposed a new test-time optimization method for estimating complete and globally consistent motion for an entire video. 
We introduced a new video motion representation called OmniMotion which includes a quasi-3D canonical volume and per-frame local-canonical bijections. OmniMotion can handle general videos with varying camera setups and scene dynamics, and produce accurate and smooth long-range motion through occlusions. Our method achieves significant improvements over prior state-of-the-art methods both qualitatively and quantitatively.

{
\paragraph{Acknowledgements.} We thank Jon Barron, Richard Tucker, Vickie Ye, Zekun Hao, Xiaowei Zhou, Steve Seitz, Brian Curless, and Richard Szeliski for their helpful input and assistance. This work was supported in part by an NVIDIA academic hardware grant and by the National Science Foundation (IIS-2008313 and IIS-2211259). Qianqian Wang was supported in part by a Google PhD Fellowship.
}

{\small
\bibliographystyle{ieee_fullname}
\bibliography{refs}
}
\clearpage
\appendix

\section{Preparing pairwise correspondences}
\label{sec: supp input motion data}
Our method uses pairwise correspondences from existing methods, such as RAFT~\cite{teed2020raft} and TAP-Net~\cite{doersch2022tap}, and consolidates them into dense, globally consistent, and accurate correspondences that span an entire video. 
As a preprocessing stage, we exhaustively compute all pairwise correspondences (i.e., between every pair of frames $i$ and $j$) and filter them using cycle consistency and appearance consistency checks.

When computing the flow field between a base frame $i$ and a target frame $j$ as $i \rightarrow j$, we always use the flow prediction for the previous target frame ($i \rightarrow j-1$) as initialization for the optical flow model (when possible). We find this improves flow predictions between distant frames. Still, the flow predictions between distant frames can contain significant errors, and therefore we filter out flow vector estimates with cycle consistency errors~(i.e., forward-backward flow consistency error) greater than 3 pixels. 

\begin{figure}[t!]
    \centering
    \includegraphics[width=\linewidth]{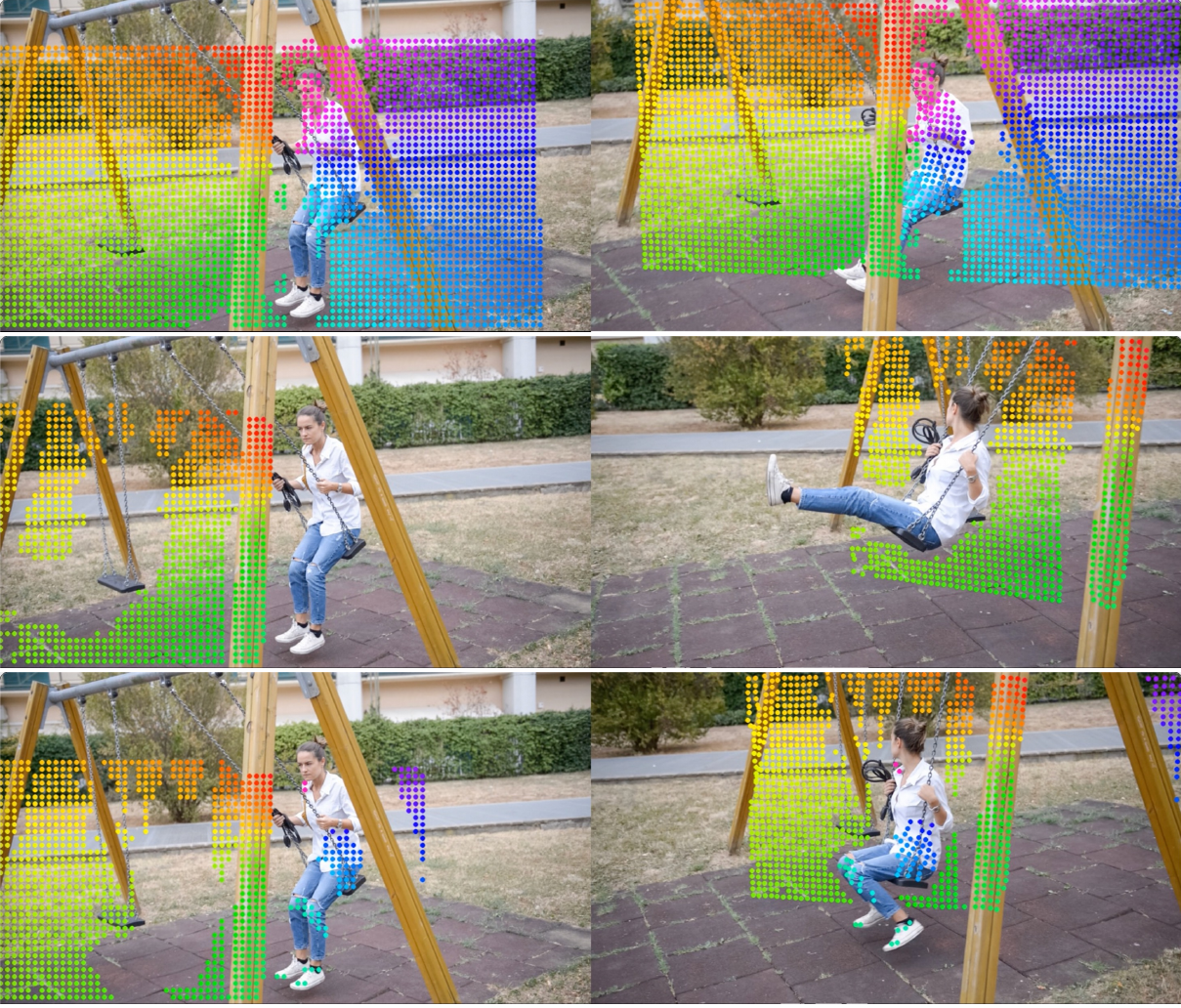}\\
    \caption{
    \small
    \textit{Correspondences from RAFT~\cite{teed2020raft} after both cycle and appearance checks}. The left column shows a single query frame, and the right column displays target frames with increasing frame distances to the query frame from top to bottom. 
    The filtered correspondences are reliable without significant errors. 
    }
    \label{fig: filtered correspondences}
\end{figure}

Despite this filtering process, we still frequently observe a persistent type of error that remains undetected by cycle consistency checks. 
This type of spurious correspondence, illustrated in Fig.~\ref{fig: cycle check}, occurs because flow networks can struggle to estimate motion for regions that undergo significant deformation between the two input frames, and instead opt to interpolate motion from the surrounding areas. In the example in Fig.~\ref{fig: cycle check}, this leads to flow on the foreground person ``locking on'' to the background layer instead. 
This behavior results in incorrect flows that survive the cycle consistency check, since they are consistent with a secondary layer's motion (e.g., background motion). 
To address this issue, we additionally use an appearance check: we extract dense features for each pixel using DINO~\cite{caron2021emerging} and filter out correspondences whose features' cosine similarity is $<0.5$. 
In practice, we apply the cycle consistency check for all pairwise flows and supplement it with an appearance check when the two frames are more than 3 frames apart. 
We found this filtering process consistently eliminates major errors in flow fields across different sequences without per-sequence tuning. The results of our filtering approach, after both cycle and appearance consistency checks, are illustrated in Fig.~\ref{fig: filtered correspondences}. 

\begin{figure}[t]
\setcounter{figure}{4}
    \centering
    \includegraphics[width=\linewidth]{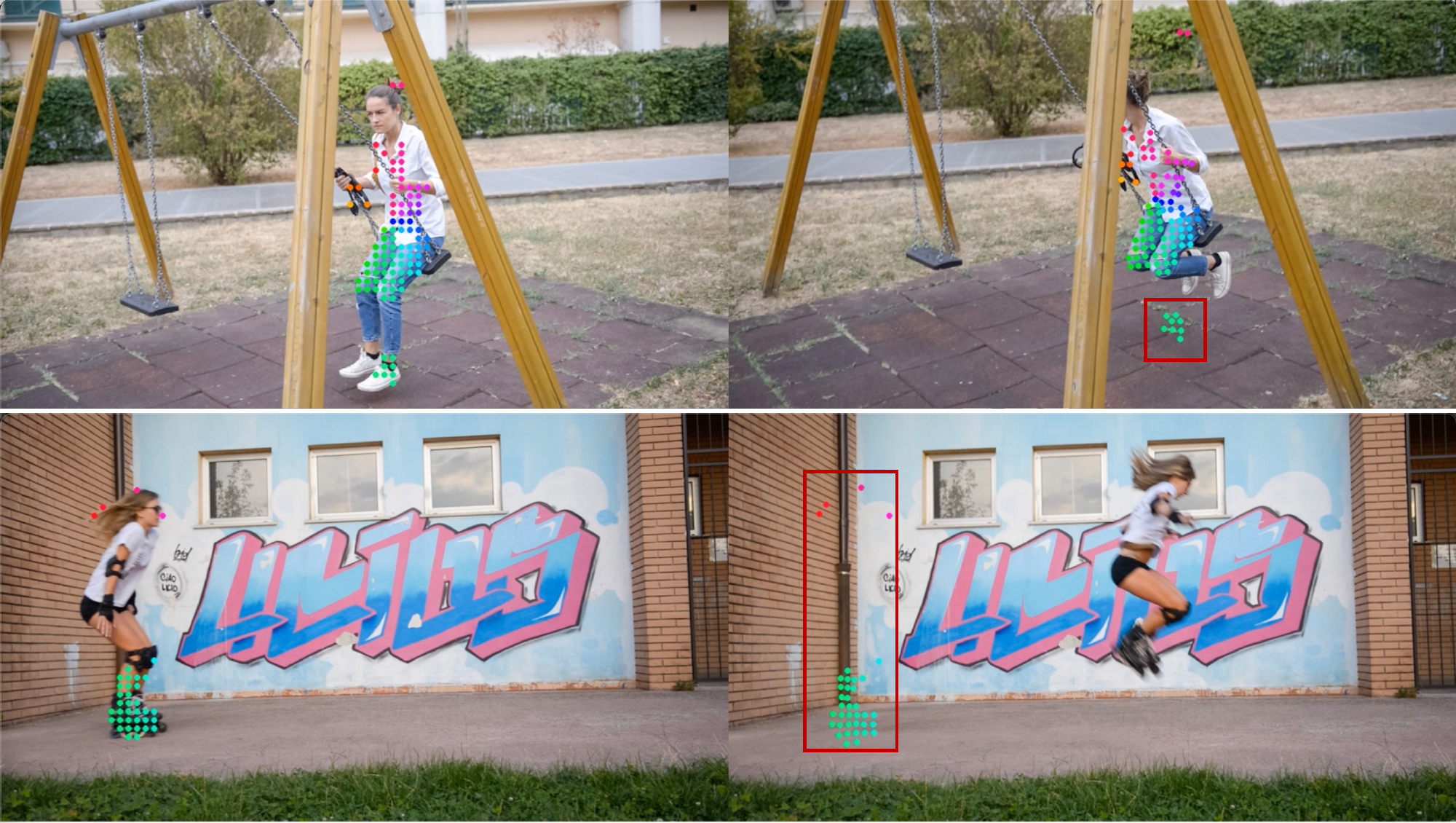}\\
    \caption{
    \small
    \textit{Erroneous correspondences after cycle consistency check}. The red bounding box highlights a common type of incorrect correspondences from flow networks like RAFT~\cite{teed2020raft} that remains undetected by cycle consistency check. 
    The left images are query frames with query points and the right images are target frames with the corresponding predictions. 
    Only correspondences on the foreground object are shown for better clarity.
    }
    \label{fig: cycle check}
\end{figure}

One drawback of such a filtering process is that it will also remove correct flows for regions that become occluded in the target frame. For certain correspondence methods (such as RAFT), including these motion signals during occlusion events can result in better final motion estimates. 
Therefore, we devise a simple strategy for detecting reliable flow in occluded regions. 
For each pixel, we compute its forward flow to a target frame~(a), cycle flow (flow back to the source frame from the target pixel)~(b), and a second forward flow~(c). 
This process effectively amounts to a 2-pass cycle consistency check: the consistency between (a) and (b) forms a standard cycle consistency check, and the consistency between (b) and (c) forms a secondary, supplementary one. We identify pixels where (a) and (b) are inconsistent but (b) and (c) are consistent and deem these to be occluded pixels. 
We found this approach effective in identifying reliable flows for occluded regions---particularly when the two frames are close to each other. 
Therefore, we allow these correspondences to bypass cycle consistency checks if they span a temporal distance of less than $3$ frames.
Our experiments use this added signal for the variant of our method that uses RAFT flow, but not for the TAP-Net variant, as we found the predicted correspondences from the latter were less reliable near occlusion events. 

We can also optionally augment the supervising input flow by chaining sequences of correspondences that are deemed reliable~(i.e., those that satisfy the cycle consistency and appearance consistency checks). This helps densify the set of correspondences, creating supervision between distant frames where the direct flow estimates were deemed unreliable and therefore discarded during filtering. We found this process to be beneficial especially for challenging sequences with rapid motion or large displacements, where optical flow estimates between non-adjacent frames are less reliable.

\section{Additional ablations}
\label{sec: supp ablations}

In addition to the ablations in the main paper, we provide the following ablations and report the results in Table~\ref{tab: ablation supp}: 
1) \emph{Plain 2D}: Rather than using a quasi-3D representation with bijections to model motion, we utilized a simple 8-layer MLP with 256 neurons that takes the query pixel location, query time, and target time as input and outputs the corresponding location in the target frame. 
Although we applied positional encoding with $8$ frequencies to the input to enable better fitting, this ablation failed to capture the holistic motion of the video, instead only capturing simpler motion patterns for the rigid background.
2) \emph{No flow loss}: we remove the flow loss and only rely on photometric information for training. We find this approach is effective only for sequences with small motion, where a photometric loss can provide useful signals to adjust motion locally.  For sequences with relatively large motion, this method fails to provide correct results. 
3) We also vary the number of samples $K$ for each ray from 32 to 16 and 8. The resulting ablations, named \emph{\#Samples K=8} and \emph{\#Samples K=16}, demonstrate that using a denser sampling strategy tends to produce better results.

\begin{table}[t]
\centering

{\resizebox{0.85\linewidth}{!}{
\small 
\begin{tabular}{lcccc}
\toprule

\textbf{Method} & AJ~$\uparrow$ & $<\delta^x_\textrm{avg}$~$\uparrow$ & OA~$\uparrow$ & TC~$\downarrow$  \\ 
\midrule
Plain 2D & $11.6$ & $19.8$ & $76.7$ & $1.25$ \\
No invertible & $12.5$ & $21.4$ & $76.5$ & $0.97$\\
No flow loss & $23.9$ & $37.3$ & $70.8$ & $1.75$\\
No photometric & $42.3$ & $58.3$ & $84.1$ & $0.83$\\
Uniform sampling & $47.8$ & $61.8$ & $83.6$ & $0.88$ \\
\#Samples K = 8 & $48.1$ & $63.5$ & $84.6$ & $0.75$ \\
\#Samples K = 16 & $49.7$ & $65.0$ & $\mathbf{85.6}$ & $0.84$ \\
\midrule
Full & $\mathbf{51.7}$ & $\mathbf{67.5}$ & $85.3$ & $\mathbf{0.74}$ \\
\bottomrule
\end{tabular}
}}%

\caption{
\small
Ablation study on DAVIS~\cite{pont20172017}.
}
\vspace{-0.1in}
\label{tab: ablation supp}
\end{table}

\section{Additional implementation details}
\label{sec: supp details}

We provide additional implementation details below and will release our code upon acceptance. 
\paragraph{Error map sampling.}
We cache the flow predictions generated by our model every 20k steps and use them to mine hard examples for effective training. Specifically, for each frame in the video sequence, we compute the optical flow between that frame and its subsequent frame, except for the final frame where we compute the flow between it and the previous frame. We then compute the $L_2$ distance between the predicted flow and supervising input flow, where each pixel in the video is now associated with a flow error. 
In each training batch, we randomly sampled half of the query pixels using weights proportional to the flow errors and the other half using uniform sampling weights.

\paragraph{Training details.} 
In addition to the photometric loss $\losspho$ introduced in the main paper, we include an auxiliary loss term that supervises the \emph{relative} color between a pair of pixels in a frame:
\begin{equation}
    \mathcal{L}_\text{pgrad} = \sum_{    \Omega_p}||(\predcolor_i(\bm{p}_1) - \predcolor_i(\bm{p}_2)) - ( \gtcolor_i(\bm{p}_1) - \gtcolor_i(\bm{p}_2))||_1
\end{equation}
Here, $(\predcolor_i(\bm{p}_1) - \predcolor_i(\bm{p}_2))$ is the difference in predicted color between a pair of pixels, and $( \gtcolor_i(\bm{p}_1) - \gtcolor_i(\bm{p}_2))$ is the corresponding difference between ground-truth observations. This loss is akin to spatial smoothness regularizations or gradient losses that supplement pixel reconstruction losses in prior work~\cite{kasten2021layered,ranftl2020towards}, but instead computed between pairs of randomly sampled, potentially distant pixels $\bm p_1$ and $\bm p_2$, rather than between adjacent pixels.
We apply the same gradient loss to the flow prediction as well. We found that including these gradient losses helps improve the spatial consistency of estimates, and more generally improves the training process. We also use distortion loss introduced in mip-NeRF 360~\cite{barron2022mip} to suppress floaters. 

We train our network with the Adam optimizer with base learning rates of $3\times 10^{-4}$, $1\times 10^{-4}$, and $1\times 10^{-3}$ for the density/color network~$\nerf$, the mapping network~$\mappingnet$, and  the MLP that computes the latent code, respectively. We decrease the learning rate by a factor of 0.5 every $20$k step. To select correspondences during training, we begin by sampling correspondences from pairs of frames with a maximum interval of $20$, and gradually increase the window size during training. Specifically, we expand the window by one every 2k steps. 

In our loss formulation, we compute the flow loss $\lossflow$ as a weighted sum of the mean absolute error (MAE) between each pair of correspondences in a training batch. 
The weight is determined by the frame interval, and is given by $w = 1/\cos(\Delta / N' \cdot \pi / 2)$, where $\Delta$ is the frame interval, and $N'$ is the current window size. 
The coefficient $\lambda_\text{pho}$ for the photometric loss initially starts at $0$ and linearly increases to $10$ over the first $50$k steps of training. After $50$k steps, $\lambda_\text{pho}$ stays fixed at $10$. This design is motivated by our observation that the photometric loss is not effective in fixing large motion errors early on in the training process, but is effective in refining the motion. The coefficient $\lambda_\text{reg}$ for smoothness regularization is set to $20$. We use the same set of network architecture and training hyperparameters when evaluating different datasets in the TAP-Net benchmark.

When sampling on each ray, we use a stratified sampling strategy and sample $K=32$ points on each ray between the near and far depth range. Additionally, when mapping a 3D location from one local volume to another, we encourage it to be mapped within our predefined depth range to avoid degenerate solutions.

During training, we use alpha compositing to propagate the training signal to all samples along a ray. However, at inference time, we instead compute the corresponding location using the single sample with the largest alpha value, which we found to produce quantitatively similar but visually better results.

\paragraph{Network architecture for $\mappingnet$.}
\begin{figure}[t!]
    \centering
    \includegraphics[width=\linewidth]{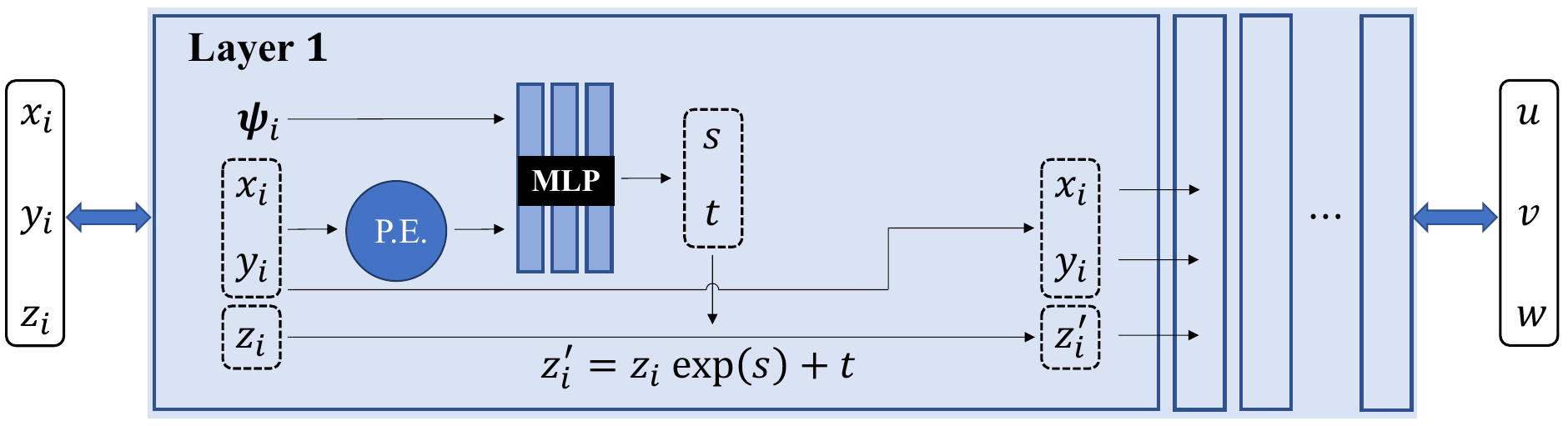}\\
    \caption{
    \small
    \textit{Network architecture for the mapping network $\mappingnet$.} We show the first affine coupling layer, which is representative of the subsequent layers, except for the different splitting patterns used. As mentioned in the main paper, this architecture is fully invertible, i.e., it can be queried in either direction, from $(u,v,w)$ to $(x,y,z)$ and vice-versa.}
    \label{fig: invertible_network}
\end{figure}
We illustrate the architecture for our invertible network $\mappingnet$ that maps between local and canonical coordinate frames in Fig.~\ref{fig: invertible_network}. $\mappingnet$ is comprised of six affine coupling layers with alternating split patterns~(only the first layer is highlighted in Fig.~\ref{fig: invertible_network}). The learnable component in each affine coupling layer is an MLP that computes a scale and a translation from a frame latent code $\latent_i$ and the first part of the input coordinates. This scale and translation is then applied to the second part of the input coordinate. This process subsequently is repeated for each of the other coordinates. 
The MLP network in each affine coupling layer has $3$ layers with $256$ channels. We found that applying positional encoding~\cite{mildenhall2021nerf} to the MLP's input coordinates improved its fitting ability, and we set the number of frequencies to 4.

\paragraph{Deformable sprites evaluation.}

Because the Deformable Sprites method defines directional mappings from image space to atlas space, we must approximate the inverses of these mappings in order to establish corresponding point estimates between pairs of frames. We do this by performing a nearest neighbor search: all points in the target frame are mapped to the atlas, and the closest atlas coordinate to the source point's mapping is chosen as the corresponding pixel. Furthermore, occlusion estimates are extracted using the following process: (1) initialize the layer assignment of source point tracks to the layer which has the higher opacity at the source frame, (2) at a given target frame index, denote the point as occluded if its originally assigned layer has lower opacity than the other layer.

\end{document}